\documentclass[AMA,STIX2COL,Linenumbersoff]{MRM}
\articletype{Journal Article}%

\received{xxx}
\revised{xxx}
\accepted{xxx}
\usepackage{pdfpages} 
\begin{document}

\title{Rician likelihood loss for quantitative MRI using self-supervised deep learning}

%\title{This is the sample article title\protect\thanks{This is an example for title footnote.}}

\author[1]{Christopher S. Parker$^1$}{\orcid{0000-0002-1616-2886}}

\author[1]{Anna Schroder$^1$}{\orcid{0009-0001-0380-0674}}

\author[1]{Sean C. Epstein$^1$}{\orcid{0000-0002-8071-1114}}

\author[1]{James Cole$^1$}{\orcid{0000-0003-1908-5588}}

\author[1]{Daniel C. Alexander$^1$}{\orcid{0000-0003-2439-350X}}

\author[1]{Hui Zhang$^1$}{\orcid{0000-0002-5426-2140}}

\authormark{PARKER \textsc{et al}}

\address[1]{\orgdiv{Centre for Medical Image Computing}, \orgname{University College London}, \orgaddress{\state{London}, \country{United Kingdom}}}

\corres{Christopher S. Parker, CMIC, 90 High Holborn, London, WC1V 6LJ. \email{christopher.parker@ucl.ac.uk}}

%\presentaddress{This is sample for present address text this is sample for present address text}

\finfo{CSP, DCA and HZ are supported by the \fundingAgency{Medical Research Council} grant \fundingNumber{MR/T046473/1}. AS is supported by the \fundingAgency{EPSRC funded i4health CDT} grant \fundingNumber{EP/S021930/1}. SCE is supported by the \fundingAgency{EPSRC-funded UCL Centre for Doctoral Training in Medical Imaging} grant \fundingNumber{EP/L016478/1}.}

%\finfo{This work was partially supported by \fundingAgency{National Science Foundation} grant \fundingNumber{DMS-2014626}}

\abstract[ABSTRACT]{
\section{Purpose} Previous quantitative MR imaging studies using self-supervised deep learning have reported biased parameter estimates at low SNR. Such systematic errors arise from the choice of Mean Squared Error (MSE) loss function for network training, which is incompatible with Rician-distributed MR magnitude signals. To address this issue, we introduce the negative log Rician likelihood (NLR) loss.
\section{Methods} A numerically stable and accurate implementation of the NLR loss was developed to estimate quantitative parameters of the apparent diffusion coefficient (ADC) model and intra-voxel incoherent motion (IVIM) model. Parameter estimation accuracy, precision and overall error were evaluated in terms of bias, variance and root mean squared error and compared against the MSE loss over a range of SNRs (5 – 30).
\section{Results} Networks trained with NLR loss show higher estimation accuracy than MSE for the ADC and IVIM diffusion coefficients as SNR decreases, with minimal loss of precision or total error. At high effective SNR (high SNR and small diffusion coefficients), both losses show comparable accuracy and precision for all parameters of both models.
\section{Conclusion} The proposed NLR loss is numerically stable and accurate across the full range of tested SNRs and improves parameter estimation accuracy of diffusion coefficients using self-supervised deep learning. We expect the development to benefit quantitative MR imaging techniques broadly, enabling more accurate parameter estimation from noisy data.}


\jnlcitation{\cname{%
\author{C.S. Parker.}, 
\author{A Schroder}, 
\author{S.C. Epstein}, 
\author{J. Cole}, 
\author{D.C. Alexander}, and
\author{H. Zhang}} (\cyear{2023}), 
\ctitle{Rician likelihood loss for quantitative MRI using self-supervised deep learning}, \cjournal{arXiv}.}

\maketitle

%\footnotetext{\textbf{Abbreviations:}~\hbox{ANA,~anti-nuclear~antibodies;~APC,~antigen-presenting~cells;} IRF, interferon regulatory factor}

%\vfill
%\pagebreak

\clearpage

\section{Introduction}\label{sec1}

% Example for\marginpar{R2.3} bibliography citations cite\cite{Taylor1937}, cites\cite{Knupp1999,Kamm2000}

Quantitative MRI (qMRI) aims to estimate and map tissue properties of interest using biophysical models that relate these properties, as parameters, to measured MR signals \cite{cercignani2018quantitative, seiberlich2020quantitative}. Such parameter estimation enables the assessment of the spatial and subject-wise variability in target tissue properties, which can inform understanding of disease processes and support objective, data-driven evaluation of patients from MRI data. For reliable inference and interpretation of tissue properties with qMRI, accurate parameter estimation is essential.

Biophysical model parameters are traditionally estimated by optimising a fitting objective function for each voxel of an MR image. This can be time-consuming, taking minutes to hours for a single volume. Furthermore, for complex non-linear models with many parameters, the objective function is often non-convex and contains local minima, increasing the propensity for estimation errors. 

Deep learning approaches, which predict the model parameters directly from measured MR data have been developed to overcome some of these limitations, demonstrating orders-of-magnitude improvement in inference speed and more robust parameter estimates \cite{golkov2016q, barbieri2020deep, kaandorp2021improved, grussu2021deep, vasylechko2022self, ottens2022deep, scalco2023quantification}. These can be divided into two categories: supervised and self-supervised. Supervised approaches aim to minimise the difference between the parameter estimates and the parameter gold standards (or ground truths) that accompany the MR data. Self-supervised approaches instead minimise a measure of loss between the signal predictions under the parameter estimates and the MR data. Supervised approaches were developed first \cite{golkov2016q}, but have been shown to produce significantly biased estimates \cite{grussu2021deep, gyori2022training}. Self-supervised approaches were in part motivated to overcome this issue \cite{barbieri2020deep}. However, studies using self-supervised learning show that bias in the estimated model parameters persists at low SNR \cite{barbieri2020deep, kaandorp2021improved, vasylechko2022self}.

The reported bias can be explained by the choice of network training loss function. So far, self-supervised networks have been trained using the Mean Squared Error (MSE) loss \cite{barbieri2020deep, kaandorp2021improved, grussu2021deep, vasylechko2022self, ottens2022deep, zhou2022unsupervised, epstein2022choice, scalco2023quantification}. This is the standard way to quantify reconstruction error in self-supervised settings \cite{bank2020autoencoders}. Minimising the MSE is equivalent to least squares estimation, which is unbiased only if the noise is centered on the true signal i.e., has zero mean \cite{jennrich1969asymptotic, kay1993fundamentals, tellinghuisen2008least}. However, MR magnitude signals used in qMRI follow a Rician distribution with non-zero mean noise \cite{gudbjartsson1995rician}. This violation of the least squares assumption may result in biased parameter estimates.

To address this, we propose a negative log Rician likelihood loss function for network training in self-supervised qMRI. Training then becomes analogous to maximum likelihood estimation (MLE), which is known to be asymptotically unbiased \cite{vandenbos1982handbook, sijbers1998maximum}. Although previous studies have recognised the need for a Rician likelihood-based loss when training self-supervised networks on MR magnitude images \cite{simpson2022motion, bolan2023improved}, none have so far demonstrated a numerically stable and accurate implementation.

In this study, we describe the theoretical basis for parameter estimation bias when using the MSE loss for self-supervised qMRI. We next introduce a numerically stable and accurate implementation of the negative log Rician likelihood loss. We then empirically evaluate the accuracy and precision of the proposed loss function using the apparent diffusion coefficient (ADC) model and intra-voxel incoherent motion (IVIM) model as example applications. Results show improved parameter estimation accuracy at low SNR with minimal loss of precision.

\section{Theory}\label{sec2}

\subsection{Self-supervised quantitative MRI}

qMRI estimates the parameters of a biophysical model of the MR signal from a set of measured signals acquired in an image voxel. The measured signals are acquired under varying experimental conditions to provide contrast for parameter estimation and are equal or larger in number than the set of model parameters.

In self-supervised qMRI, a neural network is trained to predict parameters directly from a set of signal measures by minimising a measure of error between the noise-free signals predicted by the model parameters and the measured signals. This architecture mimics that of an autoencoder \cite{bank2020autoencoders}, with the input being the measured signals, the encoder being a fully connected neural network, the latent space being the model parameters, and the decoder being the biophysical model (Fig. 1). 

\begin{figure*}[t]
\centerline{\includegraphics[width=42pc,height=16pc,keepaspectratio]{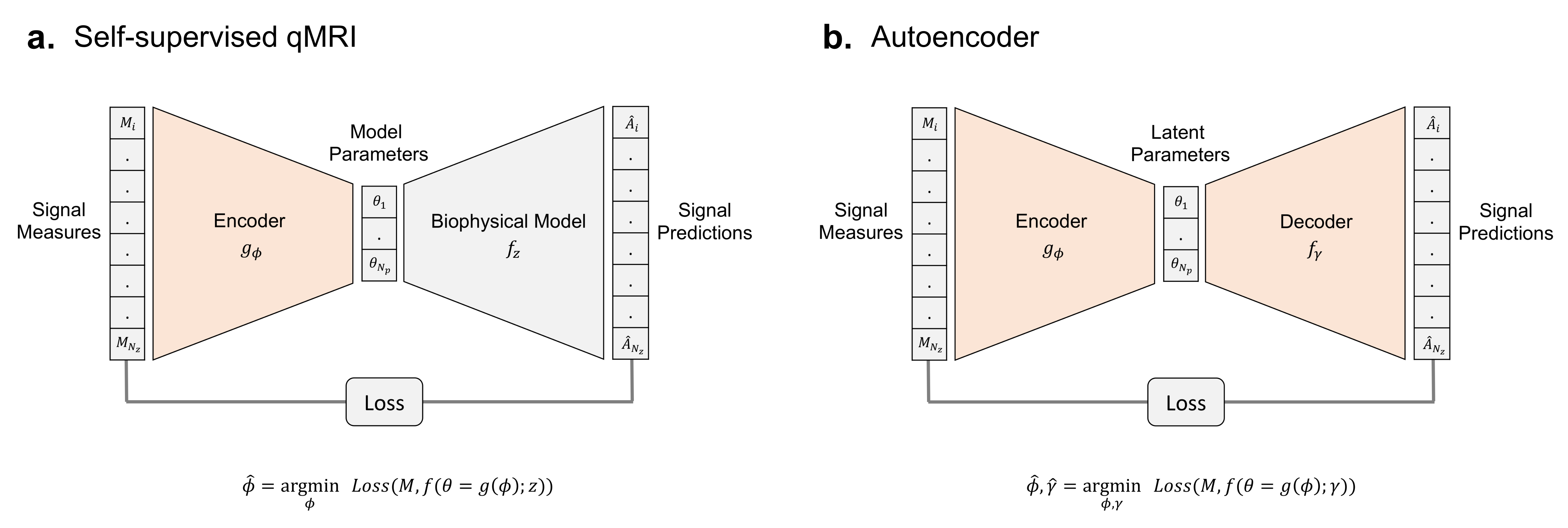}}
\caption{Comparison between self-supervised qMRI and autoencoders in terms of network architecture and optimisation. The shaded orange region denotes the trained component of the architecture. \textbf{a.} Self-supervised qMRI. The encoder network parameters are trained to output the model parameters that minimise the loss between the signal measures and the signal predictions. Signal predictions are made under the model parameters and the corresponding biophysical model. \textbf{b.} Autoencoders. The encoder and decoder network parameters are trained to minimise the loss between the signal measures and the signal predictions. Signal predictions are made under the encoded latent space parameters via the decoder.}\label{fig1}
\end{figure*}

% For displaying across multiple columns:
% https://tex.stackexchange.com/questions/30985/displaying-a-wide-figure-in-a-two-column-document

% \keepaspectratio

% /Users/christopherparker/Documents/Projects/deep_qmri/Paper/Submission/MRM/manuscript_latex
%\centerline{\includegraphics[width=16pc,height=15pc,draft]{1.png}}

\subsection{Mean squared error loss function}

Self-supervised neural networks are trained to minimise an amortised loss, which is the average of some measure of error between signal measures and signal predictions over a set of voxels. The MSE loss function, denoted $L_{MSE}$, trains networks to minimise the mean squared error between the predicted and measured MR signals, and is calculated as:

\begin{equation}
L_{MSE} = \frac{1}{N} \sum_{j=1}^{N} \left[ \frac{1}{N_z} \sum_{i=1}^{N_z} \left(M_{i,j} - \hat{A}_{i,j}\right)^2 \right]
\end{equation}

\vspace{0.5em}
\noindent where $M$ is the measured signal, $\hat{A}$ is the predicted signal, $N_z$ is the number of measured signals in a voxel and $N$ is the number of voxels in the batch. The set of signal measures in a voxel, $\{M_i \, | \, i=1,...,N_z\}$, are acquired under varying experimental conditions $\{z_i \, | \, i=1,...,N_z\}$ and the set of corresponding signal predictions, $\{\hat{A}_{i} = f \left( \theta;z_i \right) \, | \, i=1,...,N_z\}$, are generated from the chosen biophysical model with model parameters $\theta$.

The global minimum of the MSE loss function occurs when the least squares estimate of the signal predictions are attained for each voxel. Least squares predictions closely match the measured data and are unbiased estimates of the noise-free signal when the noise distribution has zero mean i.e., is centered on the noise-free signal \cite{jennrich1969asymptotic, kay1993fundamentals, tellinghuisen2008least}. Hence, least squares predictions are maximum likelihood estimates under the assumption of zero-mean Gaussian noise \cite{murphy2022probabilistic}. However, in qMRI, magnitude signal measures are typically used, which are neither Gaussian nor centered on the noise-free signal. 

%\begin{equation}
%s(nT_{s}) = s(t)\times \sum\limits_{n=0}^{N-1} \delta (t-nT_{s}) \xleftrightarrow{\mathrm{DFT}}  S \left(\frac{m}{NT_{s}}\right)
%\end{equation}

\subsection{MR magnitude signals}

The raw MR signal consists of a real and imaginary component, each of which is corrupted by zero-mean Gaussian noise. Magnitude signals are commonly calculated for quantitative MR imaging as they avoid phase artifacts. The magnitude of the complex-valued signal, $M=\sqrt{N_{R}^2 + N_{I}^2}$, where $N_R$ and $N_I$ and are real Gaussian-distributed random variables representing the real and imaginary components, follows a Rician distribution \cite{gudbjartsson1995rician} with probability density function:

\begin{equation}
p\left(M|A,\sigma \right)=\frac{M}{\sigma^2}e^{{\frac{- \left( M^2 + A^2\right)}{2\sigma^2}}} I_{0} \left( \frac{MA}{\sigma^2}\right)
\end{equation}

\vspace{0.7em}
\noindent where $I_0$ is the modified Bessel function of the first kind with order zero, $A$ is the true noise-free signal that the biophysical model aims to predict, and $\sigma$ is the noise standard deviation. The centre of this distribution, $\mathrm{E} \left[ M | A,\sigma \right]$, corresponding to the expected least squares estimate of $A$, is always greater than $A$:

\begin{equation}
\mathrm{E}\left[ M | A, \sigma \right] = \sigma \sqrt{\frac{\pi}{2}} L_{\frac{1}{2}}\left( \frac{-A^2}{2\sigma^2} \right) > A
\end{equation}

\vspace{0.7em}
\noindent where $L_{1/2}$ is the Laguerre polynomial. The lower the SNR, the greater the difference between $A$ and the least squares estimate (Fig. 2).

% figure 2 here [t] wipes away the rician section
%\begin{figure}[b]
%\centerline{\includegraphics[width=18pc,height=18pc,keepaspectratio]{2.png}}
%\caption{The expected value of the measured signal, $\mathrm{E} \left[M | A,\sigma \right]$, as a function of SNR. $\mathrm{E}\left[M | A,\sigma \right]$ corresponds to the centre of the noise and the least squares estimate of the noise-free signal at a particular SNR. Its value increasingly overestimates $A$ as SNR decreases.}\label{fig2}
%\end{figure}

\begin{figure}[b]
\centerline{\includegraphics[width=16pc,height=16pc,keepaspectratio]{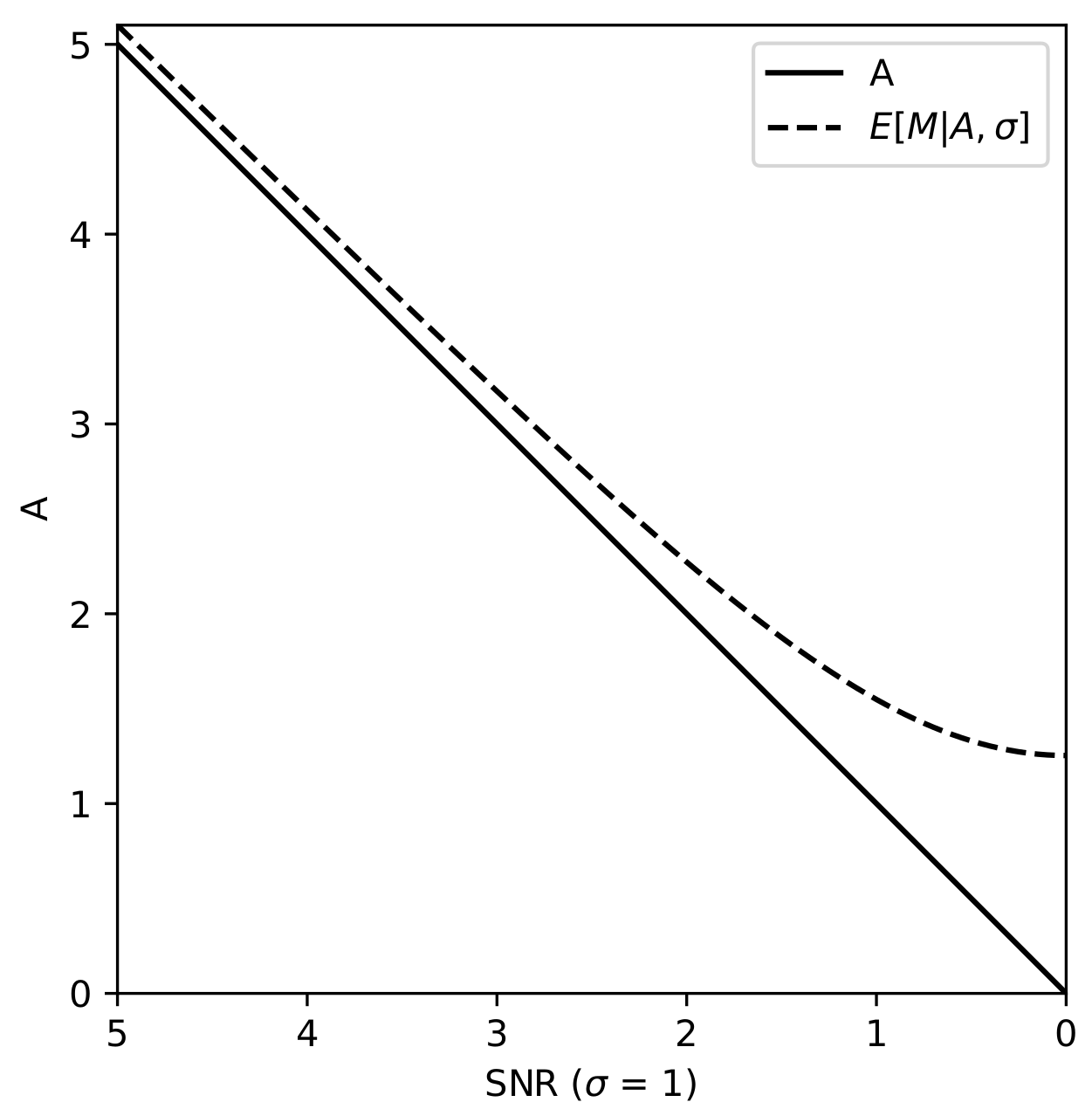}}
\caption{The expected value of the measured signal, $\mathrm{E} \left[M | A,\sigma \right]$, as a function of SNR. $\mathrm{E}\left[M | A,\sigma \right]$ corresponds to the centre of the noise and the least squares estimate of the noise-free signal at a particular SNR. Its value increasingly overestimates $A$ as SNR decreases.}\label{fig2}
\end{figure}

%\vfill\pagebreak
%\filbreak
%\newpage
%\vfill
%\pagebreak
%\clearpage

Fig. 2 and Eq. (3) show that, with sufficient degrees of freedom, signal predictions from biophysical models can lower MSE by predicting signals that are higher than the true noise-free signal. In this case the signal prediction and associated model parameter estimates are biased. Using the MSE loss for self-supervised quantitative MR imaging is therefore likely to be a source of bias in parameter estimates.

%\vspace{40em} % still some white space in column
\filbreak % works well here (but empty column created
% \bigskip does nothing
% \vspace{30em} % does something but still rician section hidden

\subsection{Rician likelihood loss function}

An alternative method for parameter estimation is MLE. MLE aims to maximise the probability of the measured data under the predicted signal and noise. This method is known to be statistically consistent, in that it is asymptotically unbiased when the model noise distribution matches that of the measured data \cite{vandenbos1982handbook, sijbers1998maximum}. An equivalent way to justify unbiasedness of MLE is that its predictive distribution is as close as possible to the empirical data distribution \cite{murphy2022probabilistic}. As MR signals are Rician-distributed, a Rician likelihood loss function may be optimised in self-supervised quantitative MR imaging to provide unbiased parameter estimates. 

The Rician likelihood of a predicted signal is given by Eq. (2). Assuming each MR measure in a voxel is independent, the total likelihood across all signal predictions for a voxel is the product of likelihoods. Network training involves minimising a loss function; in this context, Rician likelihood may be maximised by defining the training loss as the negative of the Rician likelihood. The Rician likelihood, $R$, defined for a single voxel is:
\begin{equation}
R = \prod_{i=1}^{N_z} p \left( M_i | \hat{A}_i , \sigma \right)
\end{equation}

\vspace{0.6em}
As a joint probability, Eq. (4) is subject to numerical overflow because it requires taking the product of multiple probabilities. It is common therefore to instead minimise the negative of the logarithm of the Rician likelihood. The log transformation turns multiplication into summation, reducing its propensity to overflow. Being a positively monotonic function, the log transformation preserves the location of the losses global minimum. The negative of the log Rician likelihood for a voxel is:

%\begin{equation}
%-\mathrm{log} \left( R \right) = - \sum_{i=1}^{N_z} \mathrm{log} \left( \frac{M_{i,j}}{\sigma^2} \right) - \frac{M_{i,j}^2 + \hat{A}_{i,j}^2}{2 \sigma^2} + \mathrm{log} \left( I_0 \left( \frac{M_{i,j}\hat{A}_{i,j}}{\sigma^2} \right) \right)
%\end{equation}

\begin{multline}
-\mathrm{log} \left( R \right) = - \sum_{i=1}^{N_z} \mathrm{log} \left( \frac{M_{i,j}}{\sigma^2} \right) - \frac{M_{i,j}^2 + \hat{A}_{i,j}^2}{2 \sigma^2} \\ + \mathrm{log} \left( I_0 \left( \frac{M_{i,j}\hat{A}_{i,j}}{\sigma^2} \right) \right)
\end{multline}

\subsection{Loss function specification - practical challenges}

Implementing Eq. (5) in practice is non-trivial. A naïve implementation, which directly programmes the equation as specified, has been found to be numerically unstable \cite{bolan2023improved}. Our own evaluation suggests this is because $\mathrm{log}\left( I_0  \left( x \right)  \right)$ and its gradient both contain modified Bessel functions, which grow exponentially with the input value.

To overcome this, several approaches that avoid the direct computation of $I_0  \left( x \right)$ have been reported \cite{abramowitz1972bessel, andersson2008maximum, simpson2022motion} (Table S1). We have found that none so far provide satisfactory performance in terms of both accuracy and stability across a wide range of SNRs (Fig. S1, Table S2). The approach based on log-sum-exp \cite{simpson2022motion} has poor accuracy at high SNR; this may be improved but at the cost of significantly longer computation time. Hankel’s and related approximations \cite{abramowitz1972bessel, andersson2008maximum} have poor accuracy at low SNR. Furthermore, we have found that both are subject to overflow.

We propose a new approach that is numerically stable and highly accurate for a wide range of SNRs and can be readily implemented in common machine learning packages (Table S1, Proposed). The approach utilises the exponentially scaled Bessel function $I_{0}^e \left(x \right)$ \cite{cephes:netlib, amos1985subroutine}, which approximates $I_0  \left( x \right)$ as a weighted sum of Chebyshev polynomials, $T_i$, multiplied by $e^x$ (as introduced by \citet{blair1974stable, blair1974rational}):

\begin{widetext}
\begin{equation}
I_{0}^e \left( x \right) = e^{-x} I_0 \left( x\right) :=
\begin{cases}
      e^{-x} \left[ e^{x} P_{L} \left( x \right) \right] = P_{L} \left( x \right) = \sum_{i=0}^{30} c_{i}^L T_{i}\left( x_t \right), & \text{if}\ 0 \le x \le 8 \\
      e^{-x} \left[ e^{x} P_{H} \left( x \right) / \sqrt{x} \right] = P_{H} \left( x \right) / \sqrt{x} = \sum_{i=0}^{25} c_{i}^H T_{i}\left( x_t \right) / \sqrt{x}, & \text{if}\ 8 < x < \infty
    \end{cases}
\end{equation}
\vspace{0.5em}
\end{widetext}

The coefficients are tailored for high accuracy over low and high input ranges. $x_t$ is $x$ transformed to the range $\left[-1,1 \right]$ over which the Chebyshev polynomials are defined. The transformations and coefficients are given in www.netlib.org/cephes \cite{cephes:netlib}. 

As shown in Eq. (6), the exponentially scaled Bessel function is simply the $I_0\left(x\right)$ approximation with the exponential term removed. As $ \mathrm{log} \left( I_{0}^e \left( x \right) \right) = \mathrm{log} \left( I_{0} \left( x \right) \right)-x $, then a numerically stable computation of the log Bessel that avoids direct computation of $I_{0} \left( x \right) $ is $ \mathrm{log} \left( I_{0} \left( x \right) \right) = \mathrm{log} \left( I_{0}^e \left( x \right) \right)+x $. This formulation can be readily implemented in common machine learning packages, for example in PyTorch using \textit{torch.special.i0e}. The final implementation of the negative log Rician likelihood loss function (NLR) is:

% Multi-line technique (but does not work with bracket)
% Works but cross multiple columns (Widetext does not work)
%\begin{multline}
%L_{NLR}=\frac{1}{N} \sum_{j=1}^{N} [ - \sum_{i=1}^{N_z}  \mathrm{log} \left( \frac{M_{i,j}}{2\sigma^2} \right) - \frac{M_{i,j}^{2}+\hat{A}_{i,j}^{2}}{2\sigma^2} \\
%+ \mathrm{log} \left( I_{0}^{e} \left( \frac{M_{i,j}\hat{A}_{i,j}}{\sigma^2} \right)  \right)  + \frac{M_{i,j}\hat{A}_{i,j}}{\sigma^2} ]
%\end{multline}

% vphantom technique (see Refs below)
\begin{equation}
\begin{split}
L_{NLR} &= \frac{1}{N} \sum_{j=1}^{N} \left[ - \sum_{i=1}^{N_z}  \mathrm{log} \left( \frac{M_{i,j}}{2\sigma^2} \right) - \frac{M_{i,j}^{2}+\hat{A}_{i,j}^{2}}{2\sigma^2} \right. \\
  & \left. {}+ \mathrm{log} \left( I_{0}^{e} \left( \frac{M_{i,j}\hat{A}_{i,j}}{\sigma^2} \right)  \right)  + \frac{M_{i,j}\hat{A}_{i,j}}{\sigma^2} \vphantom{\sum_{i=1}^{N_z}}\right]
\end{split}
\end{equation}

% Multi-line equation example
%\begin{multline}
 % Q(\lambda,\hat{\lambda}) = -\frac{1}{2} P(O \mid \lambda ) \sum_s \sum_m \sum_t \gamma_m^{(s)} (t) \biggl( n \log(2 \pi ) \\
  %+ \log \left| C_m^{(s)} \right| + \left( \mathbf{o}_t - \hat{\mu}_m^{(s)} \right) ^T C_m^{(s)-1} \left(\mathbf{o}_t - \hat{\mu}_m^{(s)}\right) \biggr)
%\end{multline}

% equation with left and right brackets (that would span multiple lines)
%L_{NR}=\frac{1}{N} \sum_{j=1}^{N} \left[ - \sum_{i=1}^{N_z}  \mathrm{log} \left( \frac{M_{i,j}}{2\sigma^2} \right) - \frac{M_{i,j}^{2}+\hat{A}_{i,j}^{2}}{2\sigma^2} \\
%+ \mathrm{log} \left( I_{0}^{e} \left( \frac{M_{i,j}\hat{A}_{i,j}}{\sigma^2} \right)  \right)  + \frac{M_{i,j}\hat{A}_{i,j}}{\sigma^2} \right]

% working equation
%L_{NR}=\frac{1}{N} \sum_{j=1}^{N} \left[  - \sum_{i=1}^{N_z}  \mathrm{log} \left( \frac{M_{i,j}}{2\sigma^2} \right) - \frac{M_{i,j}^{2}+\hat{A}_{i,j}^{2}}{2\sigma^2} + \mathrm{log} \left( I_{0}^{e} \left( \frac{M_{i,j}\hat{A}_{i,j}}{\sigma^2} \right)  \right)  + \frac{M_{i,j}\hat{A}_{i,j}}{\sigma^2} \right]

% Refs
%working example (https://tex.stackexchange.com/questions/194236/why-does-not-return-a-new-line-in-an-equation)
% https://tex.stackexchange.com/questions/3782/how-can-i-split-an-equation-over-two-or-more-lines
%\begin{equation}
%\begin{aligned}
%A_0 = \frac{1}{(\alpha+t_x)^{r+s+x}}{}_2F_1\left(r+s+x,x+1;r+s+x+1;\frac{\alpha-\beta}{\alpha + t_x} \right ) \\
 %- \frac{1}{(\alpha+T)^{r+s+x}}{}_2F_1\left(r+s+x,x+1;r+s+x+1;\frac{\alpha-\beta}{\alpha + T} \right ),
%\end{aligned}
%\end{equation}

\vspace{1em}
$L_{NLR}$’s global minimum is obtained when the MLE of the model parameters are predicted for each voxel. In this case, and under asymptotic conditions (as the number of measured data approaches infinity), the negative Rician loss function will train networks to predict unbiased parameter estimates. As it is not possible to assess unbiasedness of $L_{NLR}$ analytically for a finite number of measured signals per voxel, empirical observation using simulated data is required to assess estimation performance.

Note that the likelihood requires users to specify the value of the noise standard deviation, $\sigma$, which can be considered to be constant over voxels at a given SNR. In this work we use an unbiased method to estimate $\sigma$, although several other methods have also been proposed (see section 3.2.3).

%\begin{widetext}
%\begin{equation}
%s(nT_{s}) = s(t)\times \sum\limits_{n=0}^{N-1} \delta (t-nT_{s}) \xleftrightarrow{\mathrm{DFT}}  S \left(\frac{m}{NT_{s}}\right) = \frac{1}{N} \sum\limits_{n=0}^{N-1} \sum\limits_{k=-N/2}^{N/2-1} s_{k} e^{\mathrm{j}2\pi k\Delta fnT_{s}} e^{-j\frac{2\pi}{N}mn}
%\end{equation}
%\end{widetext}

\section{Methods}\label{sec3}

\subsection{Quantitative MR biophysical models}

We investigate parameter estimation accuracy and precision for two qMRI models: the ADC model \cite{le1988separation} and the intra-voxel incoherent motion (IVIM) model \cite{le1986mr}. 

\subsubsection{ADC model}

The ADC model is a simple mono-exponential signal decay model, a widely used model in diffusion imaging and across various qMRI applications, such as relaxometry. It also serves as a base for more complex qMRI models. 

The ADC model attempts to capture the extent of apparent water diffusion within a voxel. By assuming isotropic and Gaussian diffusion in all directions, the diffusion-weighted signal decays monoexponentially and is parameterised by the apparent diffusion coefficient $D$ and the signal in the absence of diffusion weighting, $S_0$. The predicted MR signal, $\hat{A}_{ADC}$, for the ADC model is:

\begin{equation}
\hat{A}_{ADC}=S_0 e^{-bD}
\end{equation}

\vspace{0.5em}
To estimate the ADC model parameters multiple signals are acquired in a single voxel across a range of b-values. $D$ is upper-bounded by the diffusivity of free water (3 $ \mathrm{\mu m^2 / ms} $ at body temperature), with lower values indicating the presence of microstructural barriers to water diffusion (e.g., cell membranes).

\subsubsection{IVIM model}

The IVIM model is to date the most widely used model for qMRI studies that use self-supervised deep learning.

	The IVIM model extends the ADC model to account for incoherent motion (zero-centered displacement) of water within capillaries. In the presence of capillary incoherent motion and isotropic Gaussian diffusion, the diffusion-weighted signal decays biexponentially and is parameterised by the diffusion coefficient of tissue $D_t$, pseudo-diffusion coefficient of the capillaries $D_p$, signal fraction of the pseudo-diffusion component $f$, and $S_0$ as defined above. The predicted MR signal, $\hat{A}_{IVIM}$, for the IVIM model is:

\begin{equation}
\hat{A}_{IVIM}=S_0 \left( f \cdot e^{-b \, \left( D_p + D_t \right)} + \left( 1 - f \right) \cdot e^{-b D_t} \right)
\end{equation}

\vspace{0.5em}
As with the ADC model, to estimate the IVIM parameters, multiple signals are acquired in a single voxel across a range of b-values. As incoherent motion within capillaries is faster than diffusion, lower b-values are often acquired to capture the rapid signal decay.  $D_t$ behaves similar to $D$ of the ADC model, but $D_p$, which models pseudo-diffusion due to blood flow in randomly oriented capillaries, is orders of magnitude larger than $D_t$. The fraction $f$ reflects the relative volume fraction of capillaries in a voxel.

\subsection{Self-supervised neural network}

\subsubsection{Network architecture}

A feed-forward deep neural network was constructed in PyTorch (v 1.21.1) to estimate voxel-wise qMRI model parameters using self-supervised learning. Following \citet{barbieri2020deep}, the network has one input layer, three fully connected hidden layers (the encoder), and one output layer (the model parameters). Three fully connected layers has been shown to be optimal for quantitative imaging of intra-voxel incoherent motion (IVIM) parameters \cite{barbieri2020deep}. 

The number of nodes in the input layer and in each hidden layer was equal to the number of acquired MR signals per voxel. The number of nodes in the output layer of the encoder was set equal to the number of model parameters for the corresponding quantitative MR model - two for ADC ($D$ and $S_0$), or four for IVIM ($D_t$, $D_p$, $f$ and $S_0$). Encoder nodes used an exponential linear unit activation function \cite{clevert2015fast}. The predicted MR signal was generated from the output layer values following the corresponding qMRI model, Eq. 8 for the ADC model or Eq. 9 for the IVIM model. To output signal predictions, b-values for predicted signals were ordered identically to the b-values used to simulate the input measurements. For the backpropagation to take approximately equally sized parameter update steps, b-values were specified in units of $\mathrm{ms / \mu m^2}$ so that the diffusion coefficient model parameter ($D$ for ADC and $D_t$ for IVIM) and $S_0$ were of the same magnitude (0.4-2 $\mathrm{ \mu m^2 / ms}$ and 0-1 a.u., respectively).

\subsubsection{Loss functions}

Networks were constructed with either the negative log Rician likelihood loss function, $L_{NLR}$, given in Eq. (7), or the MSE loss function, $L_{MSE}$, given in Eq. (1).

\subsubsection{Sigma estimation}

The NLR loss function requires an estimate of the noise standard deviation $\sigma$, assumed to be constant across voxels at a given SNR. In this case, $\sigma$ may be estimated from a background image region where there is no signal. In such regions, the Rician distribution becomes a Rayleigh distribution, with expected mean:

\begin{equation}
\mathrm{E} \left[ M \right] = \sigma \sqrt{ \frac{\pi}{2}}
\end{equation}

\vspace{0.7em}
\noindent Using the sample mean estimate of $\mathrm{E} \left[ M \right]$, an estimate of $\sigma$ is:

\begin{equation}
\hat{\sigma} = \frac{\sum_{i=1}^{N_{BG}}M_i}{N_{BG}\sqrt{\frac{\pi}{2}}}
\end{equation}

\vspace{0.7em}
\noindent where $N_{BG}$ is the number of background voxels, which in this study was set to 10,000. \citet{sijbers1999parameter} show that this estimate is unbiased. Other methods to estimate $\sigma$ also exist (e.g., \citet{sijbers1999parameter, sijbers2007automatic, rajan2010noise}).

\subsection{Network training}

\subsubsection{Simulated MR signals}

Simulated MR signals were used for network training. At a given SNR (for the b=0 $\mathrm{ms/ \mu m^2}$ signal), measured MR signals were simulated for 200,000 training voxels. Ground truth model parameter values for the training data were sampled uniformly from 10 equidistant numbers covering the range of physiologically plausible values: for the ADC model, [0.4,2]  $ \mathrm{\mu m^2 / ms}$ for $D$ and $\left[0.8,1.2 \right]$ a.u for $S_0$; for the IVIM model, $\left[ 0.4,2 \right]  \mathrm{\mu m^2 / ms}$ for $D_t$, $\left[10,150 \right]  \mathrm{\mu m^2/ms}$ for $D_p$, $\left[0.1,0.5 \right]$ for $f$ and $\left[0.8,1.2 \right]$ a.u. for $S_0$. For each voxel, diffusion MR signals were simulated under a range of b-values and complex Gaussian noise was added before computing the magnitude to produce Rician-distributed signal measures. For the IVIM model, ten b-values, representative of practical protocols, were chosen: 0, 10, 20, 30, 50, 80, 100, 200, 400 and 800 $\mathrm{ms/\mu s^2}$.  For the ADC model, ten b-values linearly spaced between 0 and 1000 $\mathrm{ms/\mu s^2}$ (inclusive) were chosen to provide comparable signal coverage as the IVIM model. MR signal measures for an additional 1,000 voxels were also simulated to calculate validation loss. Data were simulated for SNRs of 30, 20, 10, 7.5 and 5. This range covers the SNRs commonly observed in real data (typically 10-30) and includes even lower SNRs to test the limits of parameter estimation performance. Note hereafter that the unqualified term “SNR” will refer to the SNR of the signal at b=0 $\mathrm{ms/\mu s^2}$, whereas “effective SNR” will refer to the signal acquired at a particular b-value and set of model parameter values.

\subsubsection{Optimisation}

For each SNR, self-supervised network parameters were optimised to perform parameter estimation from the set of simulated MR training signals using either the NLR or MSE loss function. Network weights and biases were updated using stochastic gradient descent via backpropagation. An Adam optimiser was used with the default PyTorch settings (learning rate=0.001, betas=(0.9, 0.0999), weight decay=0) \cite{kingma2014adam}. Batch sizes of 256 voxels were used in each gradient descent step. To prevent dependence on the random initialisation of weights and biases, a common initialisation was used for both loss functions at each SNR. The common initialisation was determined as the trained network with lowest validation loss over 16 training repetitions using the NLR loss. Training was then initiated using the common initialisation for each loss function. Training was terminated if the loss function did not improve for 50 consecutive epochs or if a total of 300 training epochs were reached.

\subsection{Evaluation of parameter estimation performance}

For each SNR, 200,000 unseen simulated test voxels were produced identically to the training data (i.e., uniformly sampled across the parameter space) but with different noise instantiations. The network trained at the corresponding SNR then predicted the model parameter values from the test voxel signal measures and the error was calculated (estimate minus ground truth). Parameter estimation performance was evaluated in detail at one representative low and one high SNR, and then examined across a range of SNRs.

Parameter estimation performance was evaluated at SNR=10 (representing typical low SNR) and SNR=30 (representing typical high SNR). Metrics of accuracy and precision were calculated as a function of the parameter value. For each unique parameter combination, bias (the average error) against ground truth was calculated to quantify accuracy, standard deviation was calculated to quantify precision and root mean squared error (RMSE) was calculated to report overall deviation from ground truth. To visualise these metrics and their variation for a particular parameter and parameter value, the average and standard deviation of each metric was calculated over the unique parameter combinations of the remaining parameter(s) (e.g., over the unique values of $S_0$ for $D$ in the case of the ADC model). 

Performance was then studied across a range of SNRs for each model parameter to observe the overall error trends and to allow for direct comparison of performance to \citet{barbieri2020deep}. At each SNR, the distribution of all errors across all training voxels was plotted for SNRs of 30, 20, 10, 7.5 and 5.

\section{Results}\label{sec4}

\subsection{Parameter estimation performance at low SNR}

At low SNR, the MSE loss showed significant bias in diffusion coefficient estimation that worsened with higher diffusion coefficients; that is – when the effective SNR (the SNR of the diffusion-weighted signal under the model parameters) was lower; with bias reaching ~5\% and ~20\% for ADC $D$ and IVIM $D_p$ diffusion coefficients, respectively (Fig. 3). In comparison, the NLR loss showed comparatively improved accuracy of diffusion coefficient estimation for both models, with only minor overestimation of ADC diffusion coefficient by ~1\% at high diffusivities.
 
Precision decreased and overall error increased with higher diffusivity for both losses, as shown by increasing standard deviation and RMSE. Both losses performed equally well in terms of accuracy, precision and overall deviation from ground truth for the $D_p$ and $f$ IVIM parameters. Interestingly, both losses showed relatively worse accuracy for IVIM $D_p$ and $f$ than for $D_t$. 

% 50 works out but blank space in column
% 45 (both figures) works ok and with no blank space
% 47 (both figures) also works ok and with no blank space but starts to give more warnings
% 49 (both figures) also works ok and with no blank space but starts to give more warnings
% can also make them both very large at which point they get pushed to the end of the document & with no column space
\begin{figure*}[b]
\centerline{\includegraphics[width=49pc,height=49pc,keepaspectratio]{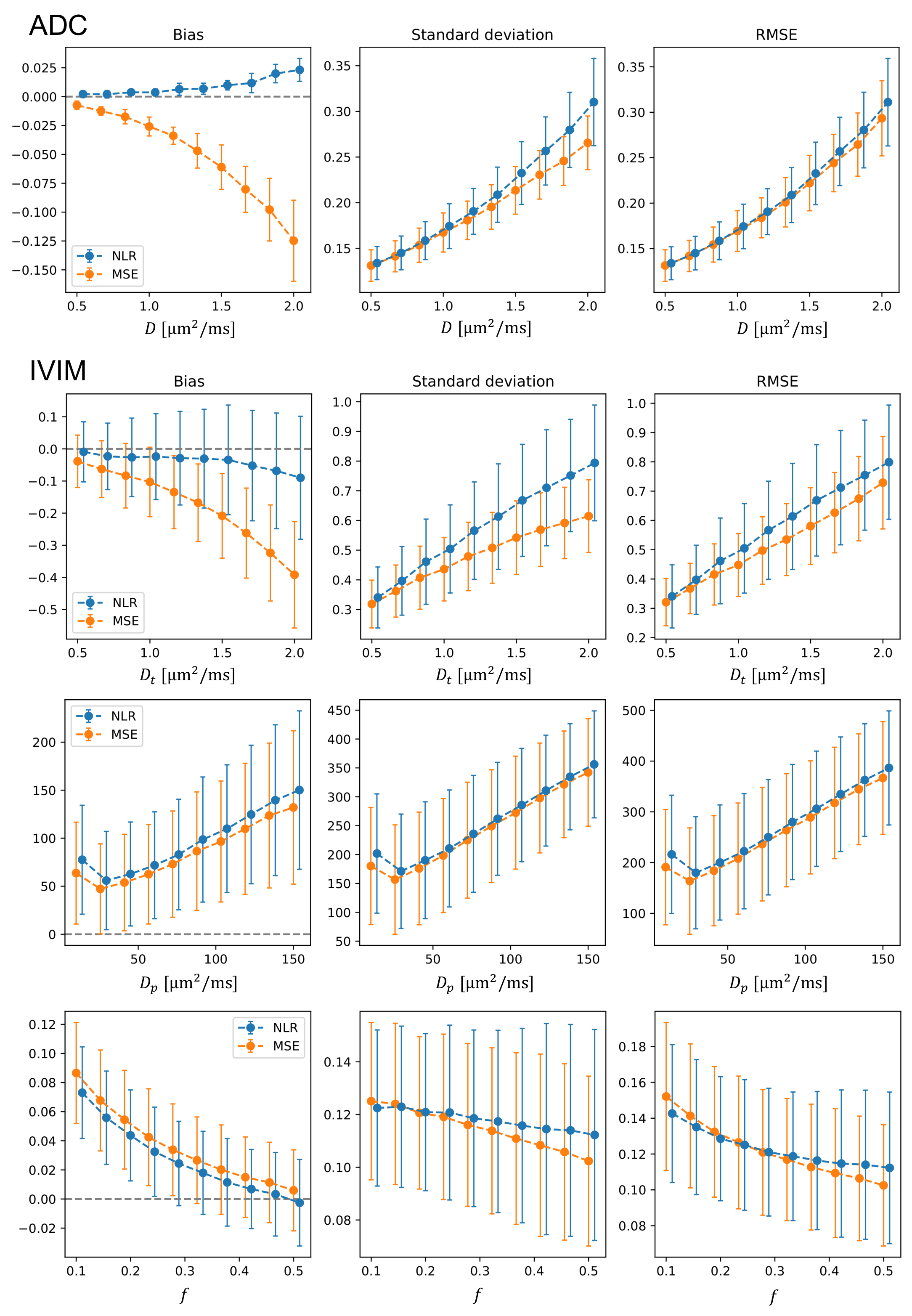}}
\caption{Comparison of estimation performance at a low SNR of 10 between self-supervised networks trained with NLR or MSE loss for ADC and IVIM model parameters. Points and error bars correspond to the mean and standard deviation of the performance metric across unique parameter values. NLR points and error bars have been offset to the right to aid visualisation.}\label{fig3}
\end{figure*}

\subsection{Parameter estimation performance at high SNR}

At high SNR, both losses showed improved accuracy and precision. However, even at high SNR, the MSE loss showed systematic underestimation of ADC diffusion coefficient $D$ as diffusivity increased (effective SNR decreased, Fig. 4, upper left), with bias reaching ~1\% at high diffusivities. In comparison, the NLR loss showed no significant bias in ADC diffusion coefficient estimation at high SNR. 
Accuracy of the IVIM model parameters $D_t$, $D_p$ and $f$ was approximately equal for MSE and NLR losses. Precision and overall deviation from ground truth was also approximately equal and generally increased with higher parameter values (lower effective SNR) for both ADC and IVIM model parameters. 

\begin{figure*}[t]
\centerline{\includegraphics[width=49pc,height=49pc,keepaspectratio]{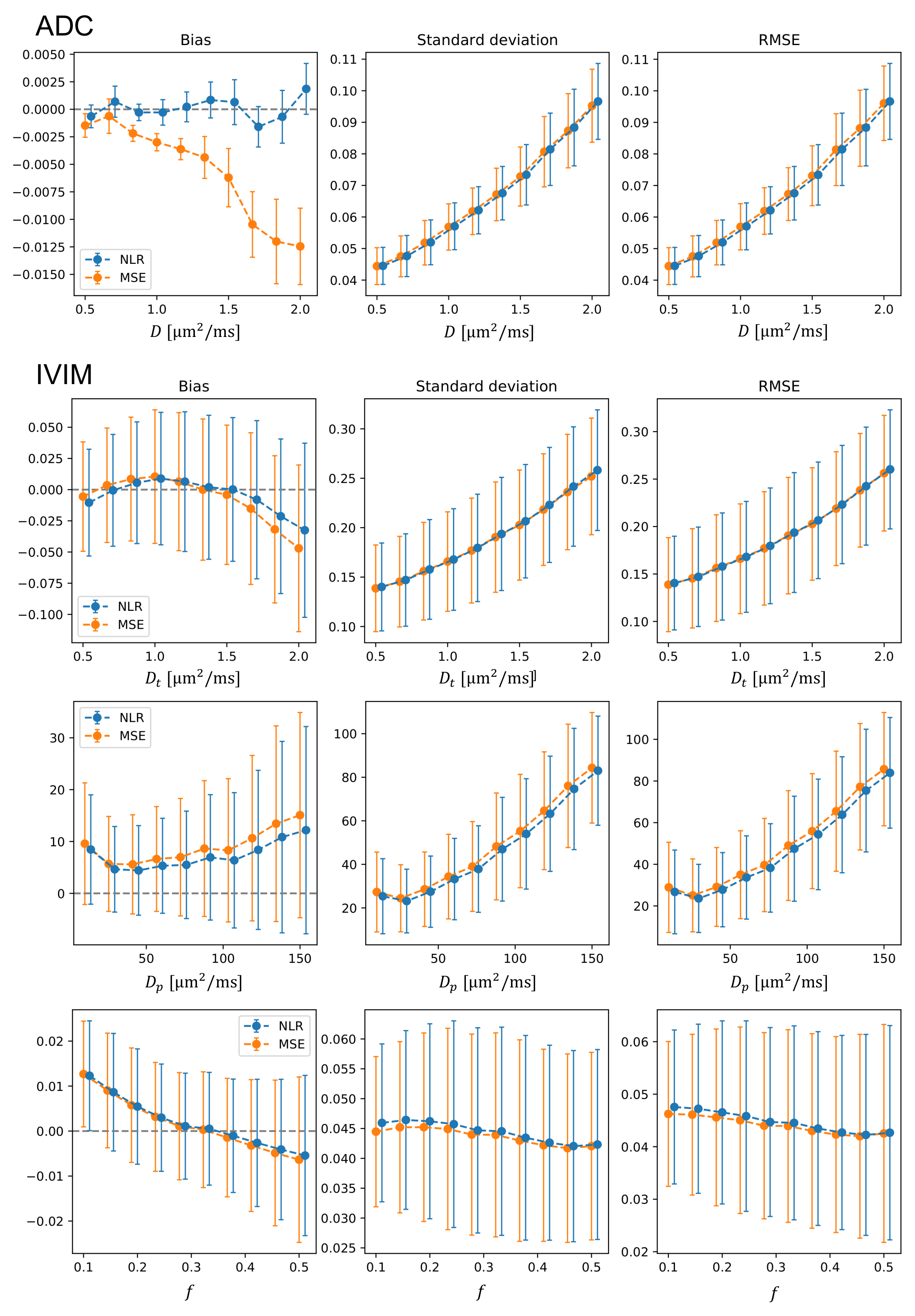}}
\caption{As in Fig. 3 but for a high SNR of 30.}\label{fig4}
\end{figure*}

\subsection{Error distribution across SNRs}

The summary of errors across SNRs shows similar trends to that observed at the representative high and low SNRs in Fig. 3 and 4. Boxplots show the MSE loss tended to underestimate the ADC and IVIM diffusion coefficient as SNR decreased (median lower than zero, Fig. 5), In comparison, the NLR loss showed a relatively more stable accuracy as SNR decreased. 

As SNR decreased, both losses showed a reduction in precision (wider interquartile range) for all metrics, with the NLR loss showing marginally lower precision than the MSE loss. Furthermore, both losses showed increasing overestimation of IVIM $f$ as SNR decreased. At high SNR (>10), both losses showed comparable error distributions.

\begin{figure*}[t]
\centerline{\includegraphics[width=37pc,height=37pc,keepaspectratio]{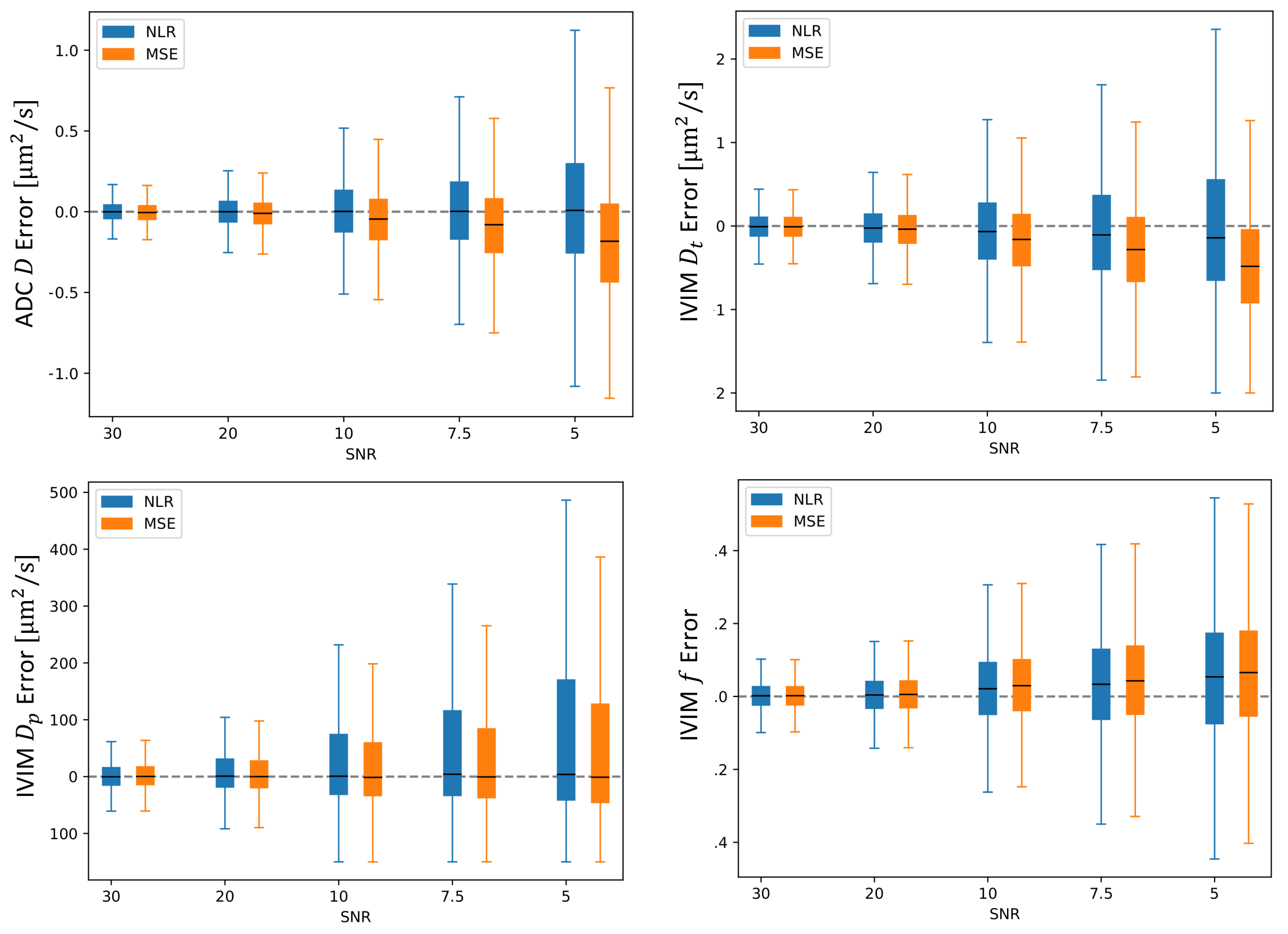}}
\caption{Boxplots of errors for ADC $D$ and IVIM $D_t$, $D_p$ and $f$ model parameters at SNRs of 30, 20, 10, 7.5 and 5. The line shows the median error across all estimates and the box shows the inter-quartile range. Whiskers extend to the most extreme data point within 1.5 times the inter-quartile range from the median.}\label{fig5}
\end{figure*}

%\begin{figure}[t]
%\centerline{\includegraphics[width=16pc,height=15pc,draft]{empty}}
%\caption{This is the sample figure caption.}\label{fig1}
%\end{figure}

\section{Discussion}\label{sec5}

Previous qMRI studies using self-supervised deep learning have demonstrated increasingly biased parameter estimates at low SNR. This work suggests that a major source of bias is the use of MSE loss function, the assumptions of which are increasingly violated as SNR decreases. To overcome this, we propose and evaluate a numerically stable negative log Rician likelihood loss function for self-supervised training. Using the ADC and IVIM models as exemplars, use of NLR loss improves estimation accuracy of diffusion coefficients as SNR decreases, with none or minimal loss of precision. 

	We propose a novel robust computational implementation of the Rician log-likelihood loss function for self-supervised training. The implementation computes the log-Bessel using the exponentially-scaled Bessel function, resulting in high numerical stability, while its approximation using Chebyshev polynomials is highly accuracy (Fig. S1). This permits calculation of the Rician log-likelihood over the full range of SNRs encountered in practical diffusion MR imaging scenarios. On the other hand, previous implementations were found to be either inaccurate or numerically unstable for these SNR ranges. Recently, \citet{simpson2022motion} proposed an approximation of the Rician log-likelihood for self-supervised learning based on finite series summation and applied this to image registration. However, as described in section 2.5, this finite series implementation is inaccurate at high SNR and was found to be numerically unstable at very low SNR (<5). 
	
Both loss functions demonstrated high estimation accuracy at high effective SNR – when the SNR was high and the model parameters resulted in relatively low signal attenuation (i.e., at small diffusion coefficients). Under these conditions, the Rician distribution is well approximated by a Gaussian and the center of the measured data, corresponding to the least squares signal prediction, is closer to the true noise-free signal. It should be noted however that even at high SNR the MSE loss can produce biased estimates when the effective SNR is low – at high SNR, as the diffusion coefficient increases, estimation accuracy becomes comparably worse for the MSE loss than the NLR loss (Fig. 4). MSE-trained networks are therefore systematically biased at low effective SNR for ADC and IVIM diffusion coefficients. 

The relatively higher accuracy of the NLR loss for diffusion coefficient estimation at low SNR (Fig. 3) is explained by its ability to account for the potential skewness of the Rician-distributed measures. Estimation of diffusion coefficients requires accurate estimation of the rate of noise-free signal decay. At high diffusivities or low SNRs, the noise-free signal, and therefore the effective SNR, is lower, resulting in signal measures that deviate increasingly upwards from the noise-free signal \cite{gudbjartsson1995rician} (Fig. 2). Under these conditions, the NLR loss maintains a relatively accurate noise-free signal prediction by optimising the probability of the data given the predicted measurement distribution. However, the MSE loss aims to reduce the residual error on the noise-free signal prediction regardless of the data distribution, ignoring the skewed distribution of measures. This results in MSE underestimating the rate of signal attenuation and underestimating the diffusion coefficient.

Underestimation of diffusion coefficient using MSE-trained networks is consistent with previous studies. Increasing underestimation of IVIM diffusion coefficient $D_t$ with lower SNR, reaching a bias of -0.4 $\mathrm{\mu m^2/ms}$ at SNR=10, broadly agrees with results from \citet{epstein2022choice, barbieri2020deep} and \citet{zhou2022unsupervised}, who report bias and median errors of -0.4 and -0.2/-0.5 $\mathrm{\mu m^2/ms}$ at low SNR, respectively. The marginally higher accuracy reported by \citet{barbieri2020deep} may be because they quantified accuracy using the median, which is more robust to skewed error distributions, while the lower accuracy reported by \citet{zhou2022unsupervised} may be due to their use of model selection which creates an additional source of variability. The ADC diffusion coefficient $D$ tends to have higher accuracy than the IVIM diffusion coefficient $D_t$, likely because ADC is a simpler model with a higher number of measurements per parameter.

For both losses, IVIM $D_p$ and $f$ parameters show overestimation that worsens at low effective SNR. This may be explained by the degenerate nature of the IVIM model:- a reduction (underestimation) of $D_t$, which leads to less predicted signal attenuation, can be offset by an increase in $D_p$ or $f$. 

The NLR loss marginally overestimated the ADC diffusion coefficient at low effective SNRs, reaching a bias of around 1\% at high SNR. This may be explained by the shallow gradient of the likelihood function at high diffusivities – the value of $D$ has less impact on the predicted signal at higher values. This means that, with finite sample size, small variation in noise instantiation can lead to relatively large increases in the MLE, thus creating a skewed and hence biased MLE distribution. Quantile-based analysis in Fig. 5 shows that the median error remains close to zero for $D$ even at very low SNR, despite the overestimation bias indicated by Fig. 3 and 4. It should also be noted that while MLE is known to be asymptotically unbiased, finite sample sizes can be a source of bias in non-linear model parameter estimation \cite{rilstone1996second}. Extending the NLR loss to account for the posterior probability of parameter estimates given a prior distribution may mitigate this problem and is the topic of current research. 

The proposed loss has wide potential application for improving the accuracy of deep-learning based qMRI, as many techniques utilise experimentally attenuated signal measures with low effective SNR. This is the case for diffusion MR imaging \cite{alexander2019imaging}, relaxometry \cite{margaret2012practical}, magnetisation transfer imaging \cite{helms2008high}, oxygen imaging \cite{christen2014tissue}, and others. In these cases, low effective SNR data can arise due to a variety of factors, including the biophysical tissue properties of imaged anatomy, the required image resolution, and the choice of acquisition parameters. Biophysical tissue properties that result in high signal attenuation will generate data with low effective SNR. An example of this is diffusion MR imaging in brain cerebrospinal fluid and along coherent white matter tracts. Higher spatial resolution is required to image small-scale anatomical features, producing data with lower SNR, for example in the spinal cord \cite{wheeler2014current} and cerebellum \cite{okugawa2005diffusion}. Furthermore, the acquisition parameters may be customised for a specific model and require highly attenuated signals, as is the case for diffusion kurtosis imaging \cite{jensen2005diffusional}. MR experimental choices aimed at increasing SNR to improve the robustness of parameter estimates using MSE-trained networks may now instead consider achieving the same accuracy using the NLR loss. 

The proposed loss also has potential application outside of qMRI. It may be utilised for machine learning applications involving MR imaging of nuclei that are relatively less abundant than protons, which therefore produce low SNR data, as is the case in sodium imaging \cite{madelin2013biomedical}. There is also potential application to prediction tasks requiring batch-based optimisation from Rician-distributed data. For example, for deep learning-based MR image de-noising and synthetic image generation.

\section{Conclusions}\label{sec6}

This work proposes and evaluates a numerically stable and accurate negative log Rician likelihood loss function for self-supervised learning in qMRI and compares its performance against the MSE. The NLR loss shows higher parameter estimation accuracy for lower effective SNR data compared to the MSE for ADC and IVIM diffusion coefficients. This has the potential to enable faster and more reliable quantitative analysis in noisy MR datasets, with broad application both within qMRI and beyond.

\filbreak
%Etiam euismod. Fusce facilisis lacinia dui. Suspendisse potenti. In mi erat, cursus id, nonummy sed, ullamcorper
%eget, sapien. Praesent pretium, magna in eleifend egestas, pede pede pretium lorem, quis consectetuer tortor sapien
%facilisis magna. Mauris quis magna varius nulla scelerisque imperdiet. Aliquam non quam. Aliquam porttitor quam
%a lacus. Praesent vel arcu ut tortor cursus volutpat. In vitae pede quis diam bibendum placerat. Fusce elementum
%convallis neque. Sed dolor orci, scelerisque ac, dapibus nec, ultricies ut, mi. Duis nec dui quis leo sagittis commodo.

\section*{Data availability}
A tutorial demonstrating the self-supervised training with negative log Rician likelihood loss, as well as PyTorch, Keras and TensorFlow implementations, is available at https://github.com/csparker/deep\_qmri.

\section*{Acknowledgments}
CSP, DCA and HZ are supported by the Medical Research Council (MR/T046473/1). AS is supported by  the EPSRC funded i4health CDT (EP/S021930/1). SCE is supported by the EPSRC-funded UCL Centre for Doctoral Training in Medical Imaging (EP/L016478/1).

%\subsection*{Author contributions}
%This is an author contribution text. This is an author contribution text. This is an author contribution text. This is an author contribution text. This is an author contribution text. This is an author contribution text. 

\subsection*{Financial disclosure}
None.

\subsection*{Conflict of interest}
The authors declare no potential conflict of interests.

%\bibliography{MRM-AMA}%
% \bibliographystyle{unsrtnat} add this (replace existing) into the .cls file to get citet working (ref below)
% https://tex.stackexchange.com/questions/174563/natbibs-citet-does-not-work
\bibliography{/Users/christopherparker/Documents/OLD_Documents/Latex/bib/pubs}%

\includepdf[pages={-}]{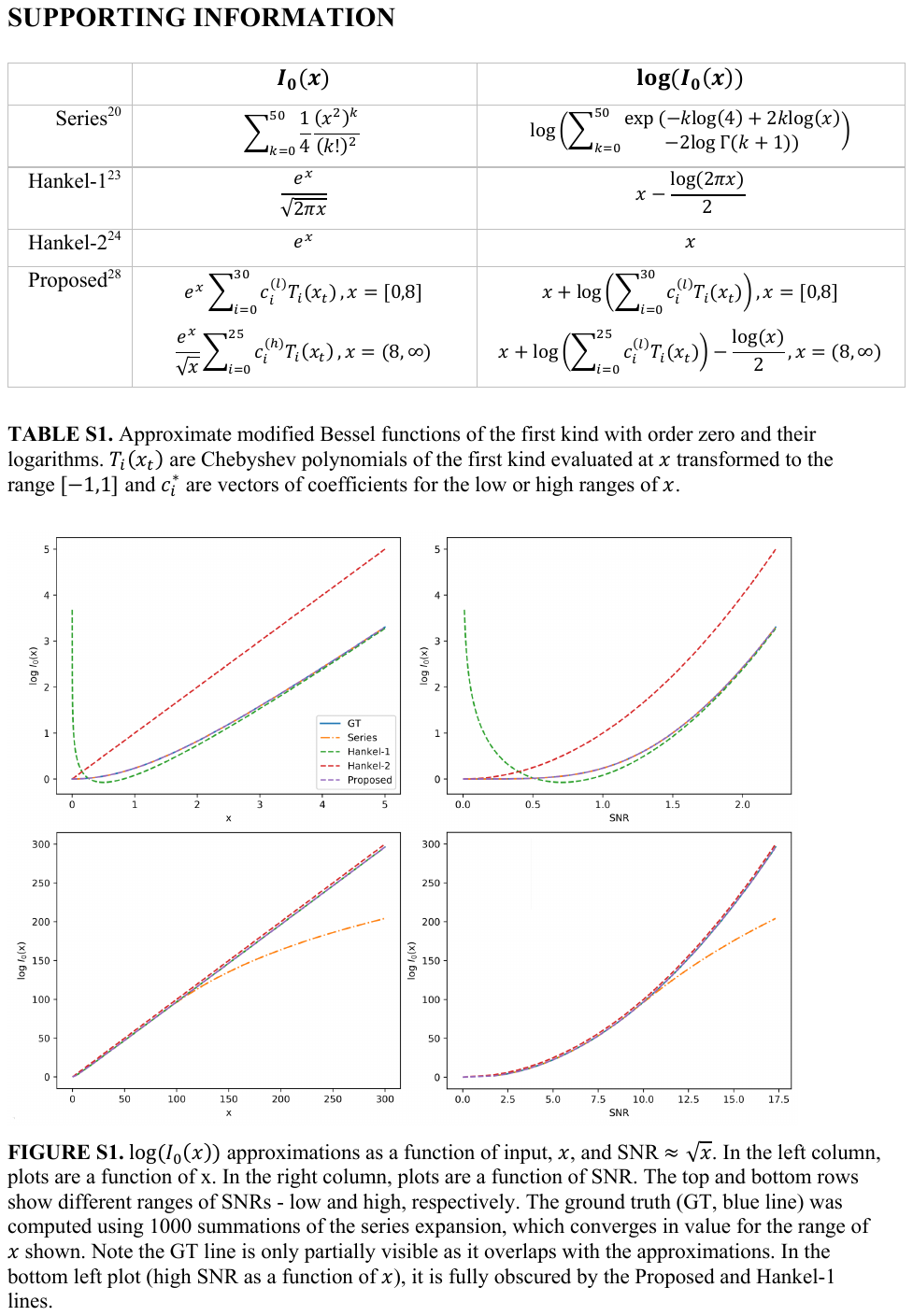}

\end{document}